\documentclass[10pt,twocolumn,letterpaper]{article}

\usepackage{cvpr}
\usepackage{times}
\usepackage{epsfig}
\usepackage{graphicx}
\usepackage{amsmath}
\usepackage{amssymb}

\usepackage[font=footnotesize,labelfont=bf]{caption}

\usepackage[belowskip=-5pt,aboveskip=3pt]{caption} 

\usepackage{multirow}
\usepackage{tabularx}
\usepackage{dsfont}
\usepackage{url}
\usepackage[tight]{subfigure}
\usepackage{color}
\usepackage{subfigure}
\usepackage{amsthm}
\usepackage[ruled,vlined]{algorithm2e}
\usepackage[export]{adjustbox}

\usepackage{comment}
\usepackage[utf8]{inputenc} 
\usepackage[T1]{fontenc}    
\usepackage{url}            
\usepackage{booktabs}       
\usepackage{amsfonts}       
\usepackage{nicefrac}       
\usepackage{microtype}      

\usepackage[titletoc]{appendix}

\definecolor{darkgreen}{rgb}{0,0.694,0.298}
\definecolor{purple}{rgb}{0.4,0.176,0.569}

\newtheorem{theorem}{Theorem}[section]
\newtheorem{lemma}[theorem]{Lemma}
\newtheorem{proposition}[theorem]{Proposition}
\newtheorem{definition}{Definition}[section]
\newtheorem{remark}{Remark}[section]

\usepackage[pagebackref=true,breaklinks=true,letterpaper=true,colorlinks,bookmarks=false]{hyperref}

\cvprfinalcopy 

\def\cvprPaperID{19} 

\ifcvprfinal\fi
\begin{document}

\setlength{\abovedisplayskip}{3pt} 
\setlength{\belowdisplayskip}{3pt} 

\title{Local Binary Convolutional Neural Networks}

\author{Felix Juefei-Xu\\
Carnegie Mellon University\\
{\tt\small felixu@cmu.edu}
\and
Vishnu Naresh Boddeti\\
Michigan State University\\
{\tt\small vishnu@msu.edu}
\and 
Marios Savvides\\
Carnegie Mellon University\\
{\tt\small msavvid@ri.cmu.edu}
}

\maketitle

\begin{abstract}
We propose \textbf{local binary convolution (LBC)}, an efficient alternative to convolutional layers in standard convolutional neural networks (CNN). The design principles of LBC are motivated by local binary patterns (LBP). The LBC layer comprises of a set of fixed sparse pre-defined binary convolutional filters that are not updated during the training process, a non-linear activation function and a set of learnable linear weights. The linear weights combine the activated filter responses to approximate the corresponding activated filter responses of a standard convolutional layer. The LBC layer affords significant parameter savings, 9x to 169x in the number of learnable parameters compared to a standard convolutional layer. Furthermore, the sparse and binary nature of the weights also results in up to 9x to 169x savings in model size compared to a standard convolutional layer. We demonstrate both theoretically and experimentally that our local binary convolution layer is a good approximation of a standard convolutional layer. Empirically, CNNs with LBC layers, called \textbf{local binary convolutional neural networks (LBCNN)}, achieves performance parity with regular CNNs on a range of visual datasets (MNIST, SVHN, CIFAR-10, and ImageNet) while enjoying significant computational savings.

\end{abstract}


\section{Introduction}\label{sec:intro}



Deep learning has been overwhelmingly successful in a broad range of applications, such as computer vision, speech recognition / natural language processing, machine translation, bio-medical data analysis, and many more. Deep convolutional neural networks (CNN), in particular, have enjoyed huge success in tackling many computer vision problems over the past few years, thanks to the tremendous development of many effective architectures, AlexNet \cite{alexNet}, VGG \cite{vgg}, Inception \cite{szegedy2015going} and ResNet \cite{resNet,identityMapping} to name a few. However, training these networks end-to-end with fully learnable convolutional kernels (as is standard practice) is (1) computationally very expensive, (2) results in large model size, both in terms of memory usage and disk space, and (3) prone to over-fitting, under limited data, due to the large number of parameters.

On the other hand, there is a growing need for deploying, both for learning and inference, these systems on resource constrained platforms like, autonomous cars, robots, smart-phones, smart cameras, smart wearable devices, \emph{etc.} To address these drawbacks, several binary versions of CNNs have been proposed \cite{binaryConnect,binaryNet,xnorNet} that approximate the dense real-valued weights with binary weights. Binary weights bear dramatic computational savings through efficient implementations of binary convolutions. Complete binarization of CNNs, though, leads to performance loss in comparison to real-valued network weights. 

In this paper, we present an alternative approach to reducing the computational complexity of CNNs while performing as well as standard CNNs. We introduce the local binary convolution (LBC) layer that approximates the non-linearly activated response of a standard convolutional layer. The LBC layer comprises of fixed sparse binary filters (called anchor weights), a non-linear activation function and a set of learnable linear weights that computes weighted combinations of the activated convolutional response maps. Learning reduces to optimizing the linear weights, as opposed to optimizing the convolutional filters. Parameter savings of at least $9\times$ to $169\times$ can be realized during the learning stage depending on the spatial dimensions of the convolutional filters ($3\times3$ to $13\times13$ sized filters respectively), as well as computational and memory savings due to the sparse nature of the binary filters. CNNs with LBC layers, called {local binary convolutional neural networks (LBCNN)}\footnote{Implementation and future updates will be available at \url{http://xujuefei.com/lbcnn}.}, have much lower model complexity and are as such less prone to over-fitting and are well suited for learning and inference of CNNs in resource-constrained environments.

Our theoretical analysis shows that the LBC layer is a good approximation for the non-linear activations of standard convolutional layers. We also demonstrate empirically that CNNs with LBC layers performs comparably to regular CNNs on a range of visual datasets (MNIST, SVHN, CIFAR-10, and ImageNet) while enjoying significant savings in terms of the number of parameters during training, computations, as well as memory requirements due to the sparse and pre-defined nature of our binary filters, in comparison to dense learnable real-valued filters.

\textbf{Related Work:}
The idea of using binary filters for convolutional layers is not new. BinaryConnect \cite{binaryConnect} has been proposed to approximate the real-valued weights in neural networks with binary weights. Given any real-valued weight, it stochastically assigns $+1$ with probability $p$ that is taken from the hard sigmoid output of the real-valued weight, and $-1$ with probability $1-p$. Weights are only binarized during the forward and backward propagation, but not during the parameter update step, in which high-precision real-valued weights are necessary for updating the weights. Therefore, BinaryConnect alternates between binarized and real-valued weights during the network training process. Building upon BinaryConnect \cite{binaryConnect}, binarized neural network (BNN) \cite{binaryNet} and quantized neural network (QNN) \cite{qnn} have been proposed, where both the weights and the activations are constrained to binary values. These approaches lead to drastic improvement in run-time efficiency by replacing most 32-bit floating point multiply-accumulations by 1-bit XNOR-count operations.

Both BinaryConnect and BNN demonstrate the efficacy of binary networks on MNIST, CIFAR-10, and SVHN dataset. Recently, XNOR-Net \cite{xnorNet} builds upon the design principles of BNN and proposes a scalable approach to learning binarized networks for large-scale image recognition tasks, demonstrating high performance on the ImageNet classification task. All the aforementioned approaches utilize high-precision real-valued weights during weight update, and achieve efficient implementations using XNOR bit count. XNOR-Net differs from BNN in the binarization method and the network architecture. In addition to network binarization, model compression and network quantization techniques \cite{squeezeNet,quantizedMobile,deepCompression,hashing,predicting,efficient,expectationBackprop,neuromorphic} are another class of techniques that seek to address the computational limitations of CNNs. However, the performance of such methods are usually upper bounded by the uncompressed and unquantized models.

Our proposed LBCNN is notably different from fully binarized neural networks and draws inspiration from \emph{local binary patterns}. LBCNN, with a hybrid combination of fixed and learnable weights offers an alternate formulation of a fully learnable convolution layer. By only considering sparse and binary weights for the fixed weights, LBCNN is also able to take advantage of all the efficiencies, both statistical and computational, afforded by sparsity and weight binarization. We demonstrate, both theoretically and empirically, that LBCNN is a very good approximation of a standard learnable convolutional layer.



\section{Forming LBP with Convolutional Filters}\label{sec:prelim}


Local binary patterns (LBP) is a simple yet very powerful hand-designed descriptor for images rooted in the face recognition community. LBP has found wide adoption in many other computer vision, pattern recognition, and image processing applications \cite{LBPbook}.

\begin{figure}
\centering
\includegraphics[width=\columnwidth]{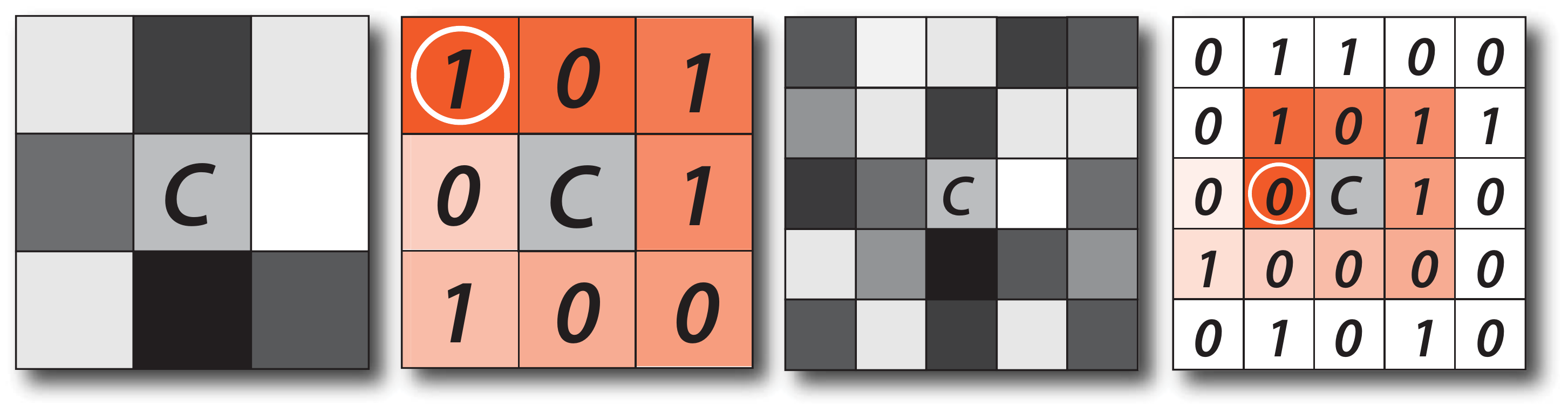}\\
\caption{(L-R) $3\times3$ patch and its LBP encoding, $5\times5$ patch and its LBP encoding.}
\label{fig:LBP_3_5}
\end{figure}

The traditional LBP operator \cite{Felix_tip14_lbp,c13,Felix_bmvc16_invert,Felix_tip15_spartans} operates on image patches of size $3\times3$, $5\times5$, \emph{etc.} The LBP descriptor is formed by sequentially compare the intensity of the neighboring pixels to that of the central pixel within the patch. Neighbors with higher intensity value, compared to the central pixel, are assigned a value of 1 and 0 otherwise. Finally, this bit string is read sequentially and mapped to a decimal number (using base 2) as the feature value assigned to the central pixel. These aggregate feature values characterize the local texture in the image. The LBP for the center pixel $(x_{c},y_{c})$ within a patch can be represented as $\mathrm{LBP}(x_{c},y_{c})=\sum^{L-1}_{n=0}s(i_{n},i_{c}) \cdot 2^{n}$
where $i_{n}$ denotes the intensity of the $n^{th}$ neighboring pixel, $i_{c}$ denotes the intensity of the central pixel, $L$ is the length of the sequence, and $s(\cdot)=1$ if $i_{n}\geq i_{c}$ and $s(\cdot)=0$ otherwise. For example, a $N\times N$ neighborhood consists of $N^{2}-1$ neighboring pixels and therefore results in a $N^{2}-1$ long bit string. Figure~\ref{fig:LBP_3_5} shows examples of LBP encoding for a local image patch of size $3\times3$ and $5\times5$.

Different parameters and configurations of the LBP formulation can result in drastically different feature descriptors. We now present a few variations that can help generalize the basic LBP descriptor:

\vspace{1mm}
\noindent\textit{\textbf{Base:}} A base of two is commonly used to encode the LBP descriptor. Consequently the weights for encoding the LBP bit string are constrained to powers of two. Relaxing these constraints and allowing the weights to take any real value can potentially generalize the LBP descriptor.

\vspace{1mm}
\noindent\textit{\textbf{Pivot:}} The physical center of the neighborhood is typically chosen as the pivot for comparing the intensity of the pixels in the patch. Choosing different locations in the patch as the pivot can enable LBP to encode different local texture patterns. Furthermore, the comparative function $s(\cdot)$ can be a function of multiple pivots resulting in a more fine-grained encoding of the local texture.

\vspace{1mm}
\noindent\textit{\textbf{Ordering:}} LBP encodes the local texture of a patch by choosing a specific order of pixels to partially preserve the spatial information of the patch. For a fixed neighborhood size and pivot, different choice of the ordering the neighbors results in different encoding of the local texture.

All the aforementioned variations \ie, the choice of pivot, the base, and the order of the encoding neighbors, are usually determined empirically and depend on the application. Being able to generalize these factors of variations in a learnable  framework is one of the motivations and inspiration behind the design of LBCNN as discussed next.

First, let us reformulate the LBP encoding more efficiently using convolutional filters. Traditional implementations of encoding LBP features use a $3\times3$ window to scan through the entire image in an overlapping fashion. At each $3\times3$ patch, the encoding involves (1) compute the difference between the pivot and the neighboring pixels (or pairs of pixels more generally), (2) a non-linear thresholding operation mapping the pixel differences to binary values, and (3) pooling the binary values through a weighed sum.
\begin{figure}
\centering
\includegraphics[width=\columnwidth]{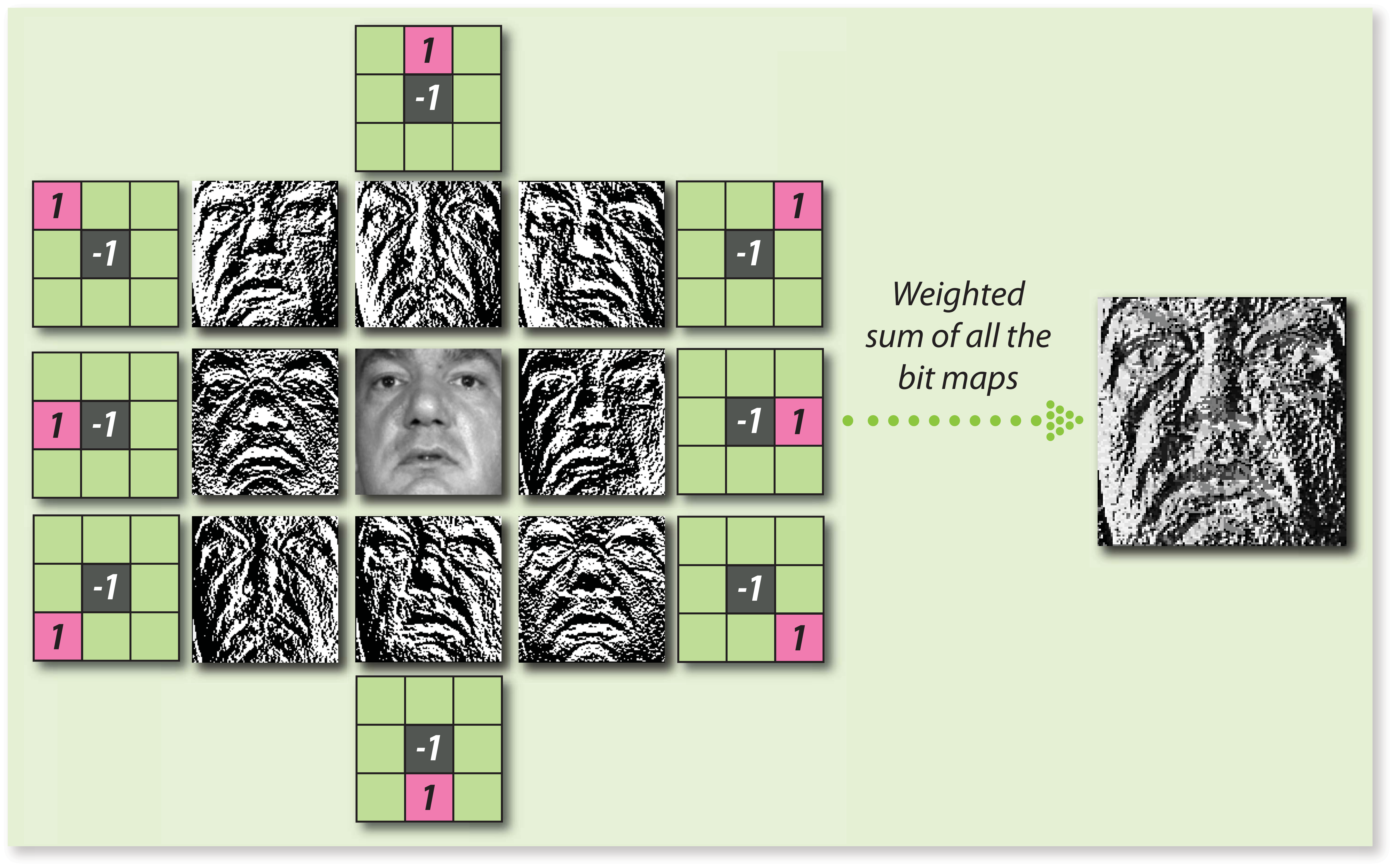}
\caption[]{Reformulation of the LBP encoding using convolutional filters.}\label{fig:WoLBP}
\end{figure}

Now, a simple convolution of the entire image with eight $3\times3$ convolutional filters, followed by simple binarization can achieve the same goal, as shown in Figure~\ref{fig:WoLBP}. Each convolution filter is a 2-sparse difference filter. The 8 resulting bit maps after binarization are also shown. Standard formulations of LBP are simply a weighted sum of all the bit maps using a pre-defined weight vector $\mathbf{v}=[2^7,2^6,2^5,2^4,2^3,2^2,2^1,2^0]$. Therefore, standard LBP feature extraction can be reformulated as $\mathbf{y} = \sum_{i=1}^8 \sigma( \mathbf{b}_i\ast \mathbf{x}  ) \cdot \mathbf{v}_i$, 
where $\mathbf{x} \in \mathds{R}^d$ is vectorized version of the original image, $\mathbf{b}_i$'s are the sparse convolutional filters, $\sigma$ is the non-linear binarization operator, the Heaviside step function in this case, and $\mathbf{y}\in \mathds{R}^d$ is the resulting LBP image.
By appropriately changing the linear weights $\mathbf{v}$, the base and the ordering of the encoding can be varied. Similarly by appropriately changing the non-zero (+1 and -1) support in the convolutional filters allows us to change the pivot. The reformulation of LBP as described above forms the basis of the proposed LBC layer.


\section{LBCNN}\label{sec:method}

\subsection{Local Binary Convolution Module}
Somewhat surprisingly, the reformulation of traditional LBP descriptor described above possess all the main components required by convolutional neural networks. For instance, in LBP, an image is first filtered by a bank of convolutional filters followed by a non-linear operation through a Heaviside step function. Finally, the resulting bit maps are linearly combined to obtain the final LBP glyph, which can serve as the input to the next layer for further processing.

\begin{figure*}
\centering
\includegraphics[width=\linewidth]{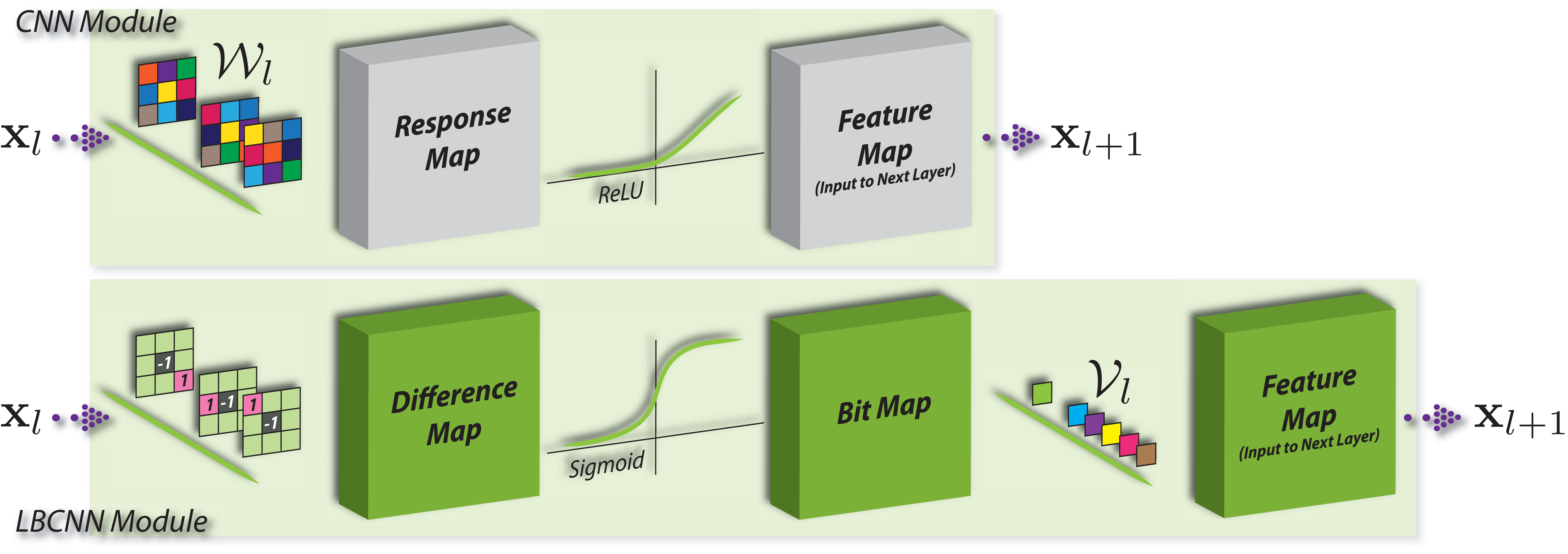}
\caption[]{Basic module in CNN and LBCNN. $\mathcal{W}_l$ and $\mathcal{V}_l$ are the learnable weights for each module.}
\label{fig:pipeline}
\end{figure*}

This alternate view of LBP motivates the design of the local binary convolution (LBC) layer as an alternative of a standard convolution layer. Through the rest of this paper neural networks with the LBC layer are referred to as local binary convolutional neural networks (LBCNN)\footnote{In this paper we assume convolutional filters do not have bias terms.}. As shown in Figure~\ref{fig:pipeline}, the basic module of LBCNN consists of $m$ pre-defined fixed convolutional filters (anchor weights) $\mathbf{b}_i, i\in[m]$. The input image $\mathbf{x}_l$ is filtered by these LBC filters to generate $m$ difference maps that are then activated through a non-linear activation function, resulting in $m$ bit maps. To allow for back propagation through the LBC layer, we replace the non-differentiable Heaviside step function in LBP by a differentiable activation function (sigmoid or ReLU). Finally, the $m$ bit maps are lineally combined by $m$ learnable weights $\mathcal{V}_{l,i},i\in[m]$ to generate one channel of the final LBC layer response. The feature map of the LBC layer serves as the input $\mathbf{x}_{l+1}$ for the next layer. The LBC layer responses to a generalized multi-channel input $\mathbf{x}_l$ can be expressed as:
\begin{align}
\mathbf{x}^t_{l+1} = \sum_{i=1}^m \sigma\left(\sum_{s} \mathbf{b}^{s}_i\ast \mathbf{x}^s_l\right) \cdot \mathcal{V}^t_{l,i}
\end{align}
\noindent where $t$ is the output channel and $s$ is the input channel. 
It is worth noting that the final step computing the weighted sum of the activations can be implemented via a convolution operation with filters of size $1\times1$. Therefore, each LBC layer consists of two convolutional layers, where the weights in the first convolutional layer are fixed and non-learnable while the weights in the second convolutional layer are learnable.

The number of learnable parameters in the LBC layer (with the $1\times1$ convolutions) are significantly less than those of a standard convolutional layer for the same size of the convolutional kernel and number of input and output channels. Let the number of input and output channels be $p$ and $q$ respectively. With a convolutional kernel of size of $h \times w$, a standard convolutional layer consists of $p\cdot h\cdot w\cdot q$ learnable parameters. The corresponding LBC layer consists of $p\cdot h\cdot w\cdot m$ fixed weights and $m\cdot q$ learnable parameters (corresponding to the $1\times 1$ convolution), where $m$ is the number of intermediate channels of the LBC layer, which is essentially the number of LBC filters. The $1\times1$ convolutions act on the $m$ activation maps of the fixed filters to generate the $q$-channel output. The ratio of the number of parameters in CNN and LBC is:
\begin{align*}
\frac{\textrm{\# param. in CNN}}{\textrm{\# param. in LBCNN}} = \frac{p \cdot h\cdot w\cdot q}{m\cdot q} = \frac{p \cdot h\cdot w}{m}
\end{align*}
For simplicity, assuming $p=m$ reduces the ratio to $h\cdot w$. Therefore, numerically, LBCNN saves at least $9\times$, $25\times$, $49\times$, $81\times$, $121\times$, and $169\times$ parameters during learning for $3\times3$, $5\times5$, $7\times7$, $9\times9$, $11\times11$, and $13\times13$ convolutional filters respectively. 

\subsection{Learning with LBC Layers}\label{sec:train}

\begin{figure}
  \centering
  \includegraphics[width=0.32\linewidth]{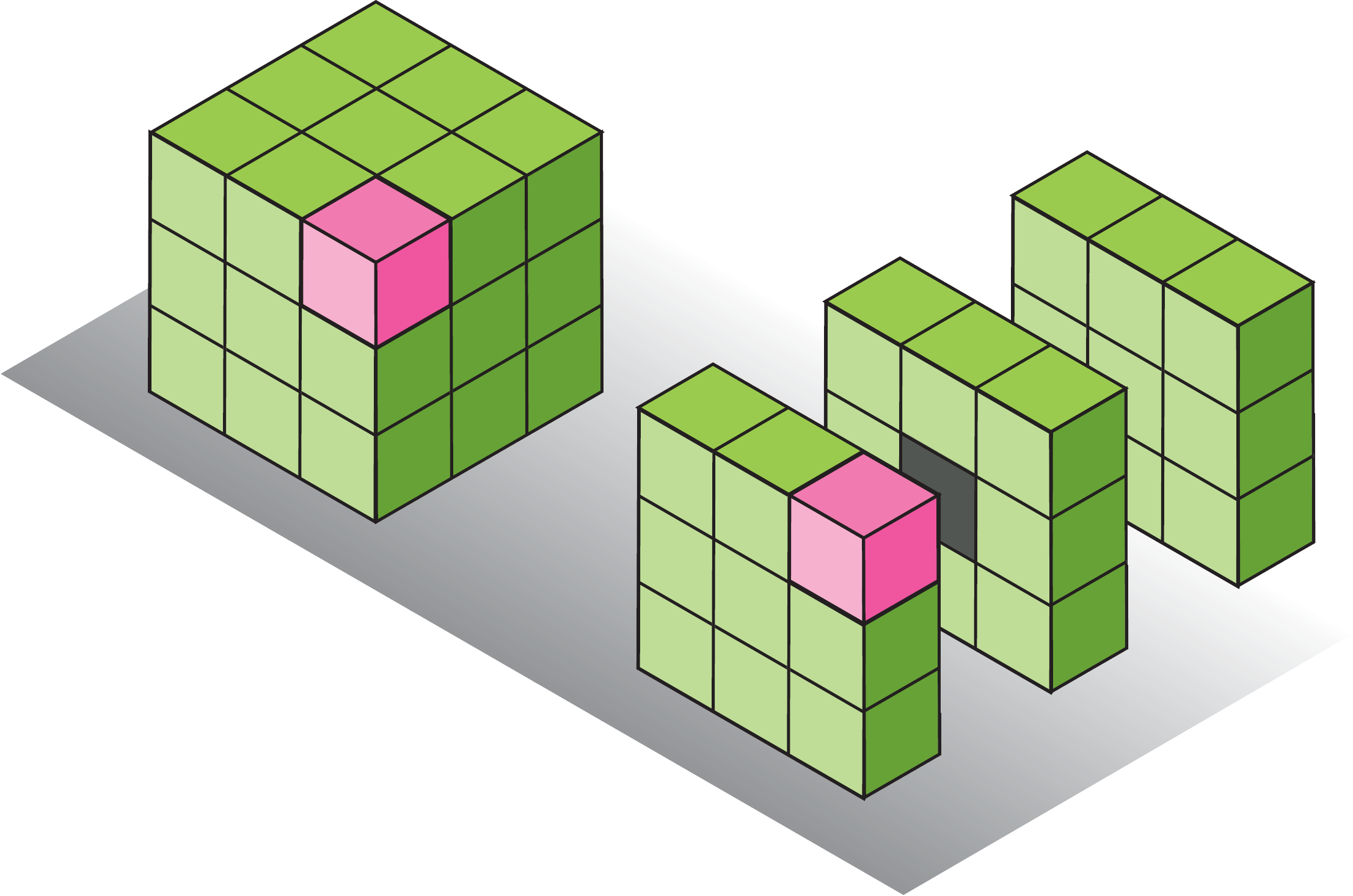}
  \includegraphics[width=0.32\linewidth]{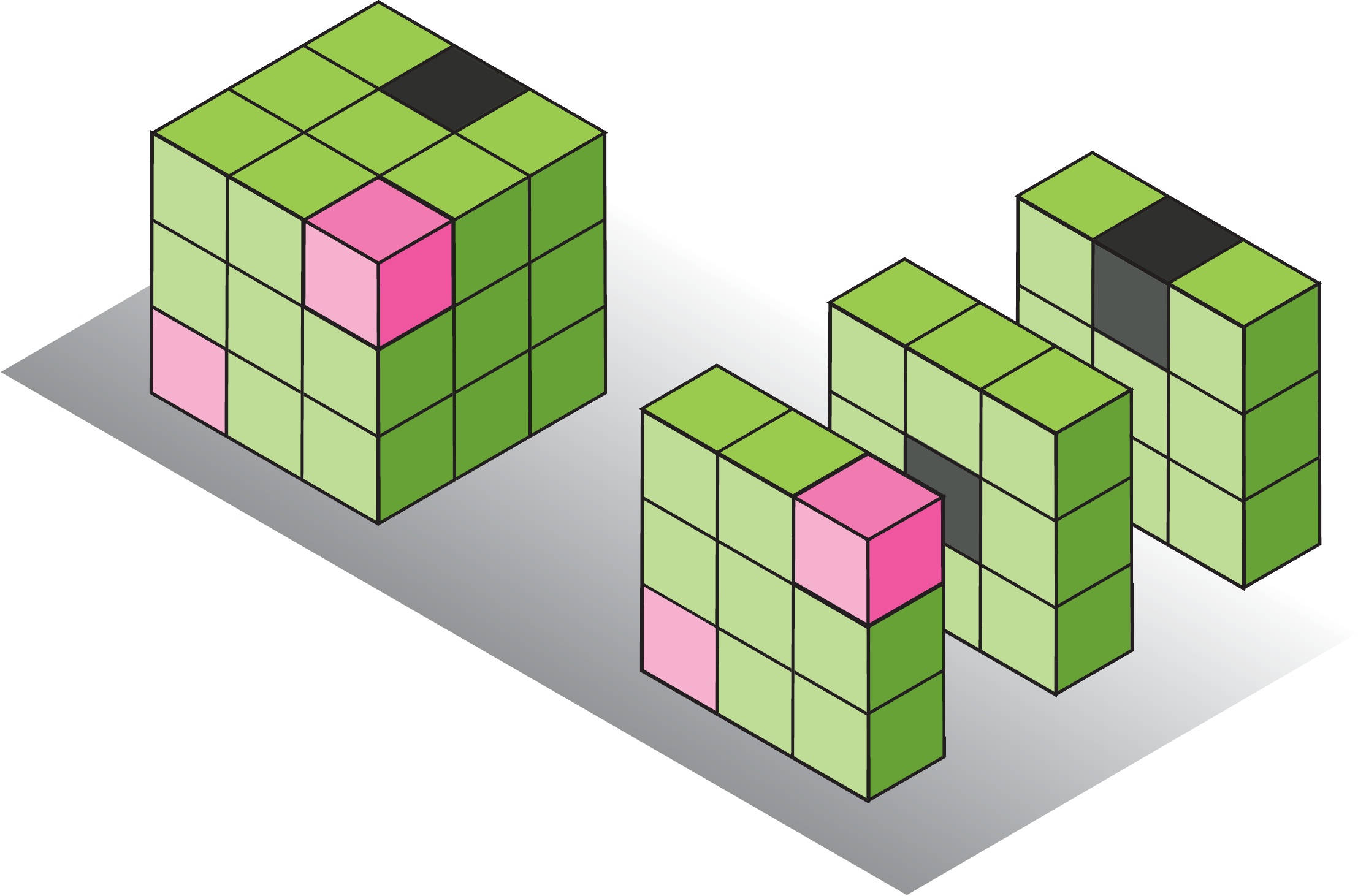}
  \includegraphics[width=0.32\linewidth]{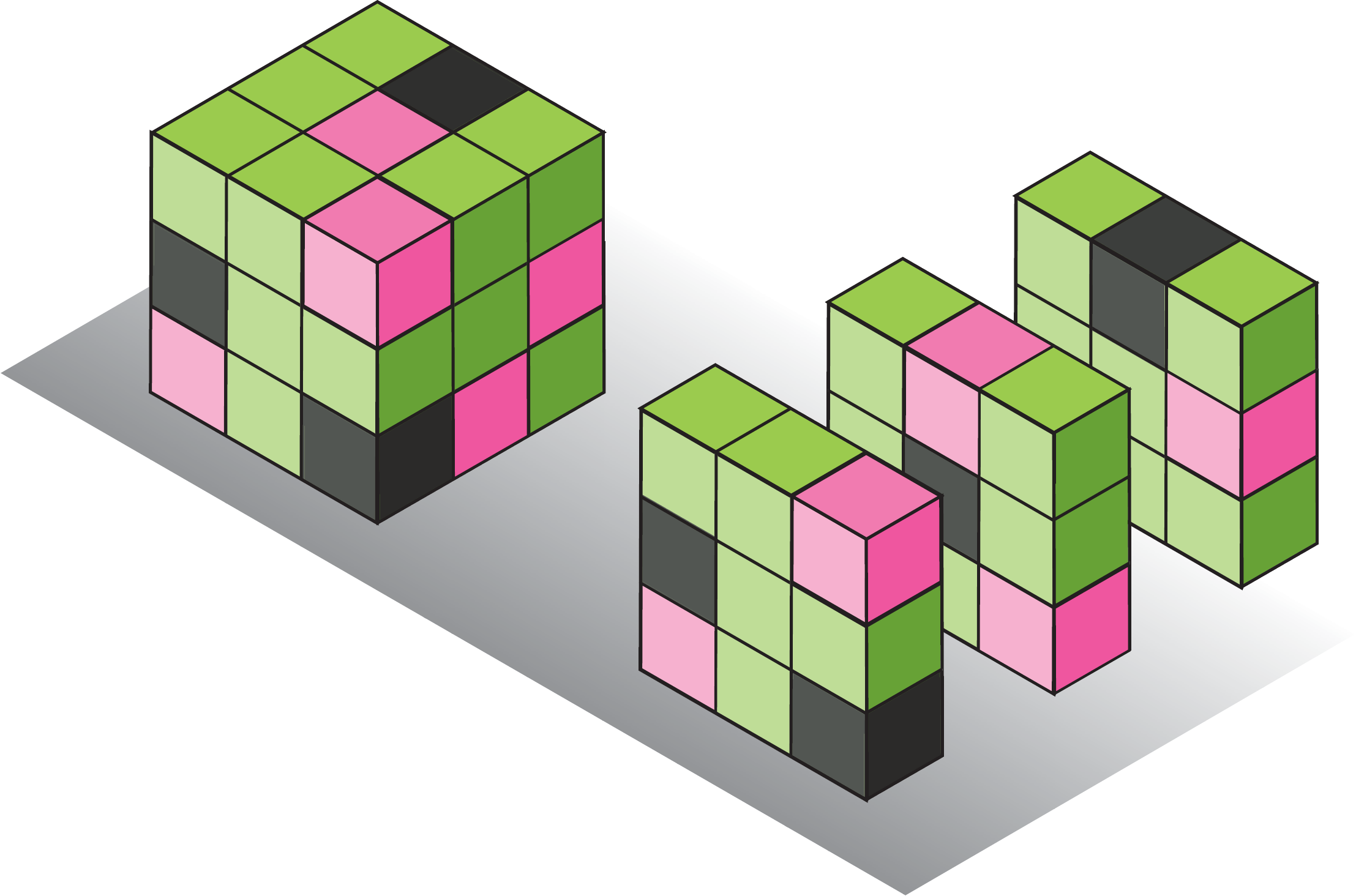}
  \caption{(L-R) Increasing sparsity level (2-sparse, 4-sparse, and 9-sparse) in the LBC filters. Pink locations bear value 1 and black locations -1. Green locations are 0. Sparsity refers to the number of non-zero elements.}
  \label{fig:sparsity}
\end{figure}

Training a network end-to-end with LBC layers instead of standard convolutional layers is straightforward. The gradients can be back propagated through the anchor weights of the LBC layer in much the same way as they can be back propagated through the learnable linear weights. This is similar to propagating gradients through layers without learnable parameters (\eg, ReLU, Max Pooling \etc). However during learning, only the learnable $1\times1$ filters are updated while the anchor weights remain unaffected. The anchor weights of size $p\times h \times w \times m$ (assuming a total of $m$ intermediate channels) in LBC can be generated either deterministically (as practiced in LBP) or stochastically. We use the latter for our experiments. Specifically, we first determine a sparsity level, in terms of percentage of the weights that can bear non-zero values, and then randomly assign 1 or -1 to these weights with equal probability (Bernoulli distribution). This procedure is a generalization of the weights in a traditional LBP since we allow multiple neighbors to be compared to multiple pivots, similar to the 3D LBP formulation for spatial-temporal applications \cite{LBPbook}. Figure~\ref{fig:sparsity} shows a pictorial depiction of the weights generated by our stochastic process for increasing (left to right) levels of sparsity\footnote{In our paper, sparsity level refers to the percentage of non-zero elements \ie, sparsity=100\% corresponds to a dense weight tensor.}. Our stochastic LBC weight generation process allows for more diversified filters at each layer while providing a fine grained control over the sparsity of the weights.

\subsection{Theoretical Analysis}

We now theoretically analyze the similarity, \ie, approximation quality, between the LBC layer and a standard convolutional layer followed by a ReLU non-linearity. We derive an upper bound on the approximation error of the LBC layer.


At layer $l$, let $\mathbf{x} \in \mathds{R}^{(p \cdot h \cdot w) \times 1}$ be a vectorized single patch from the $p$-channel input maps, where $h$ and $w$ are the spatial sizes of the convolutional filter. Let $\mathbf{w} \in \mathds{R}^{(p \cdot h \cdot w) \times 1}$ be a vectorized single convolution filter from the convolutional filter banks $\mathbf{W} \in \mathds{R}^{p\times h \times w \times m}$ with $m$ learnable  filters at layer $l$. We drop the layer subscription $l$ for brevity.

In a standard CNN, this patch $\mathbf{x}$ is projected onto the filter $\mathbf{w}$, followed by the non-linear activation resulting in the output feature value $d$. Each value of the output feature map is a direct result of convolving the input map $\mathbf{x}$ with a convolutional filter $\mathbf{w}$. This microscopic process can be expressed as:
\begin{align}
d = \sigma_{\mathrm{relu}} ( \mathbf{w}^{\top} \mathbf{x} )
\label{eq:d}
\end{align}

The corresponding output feature map value for the proposed LBC layer is a linear combination of multiple elements from the intermediate bit maps (implemented as $1\times1$ convolution). Each slice of this bit map is obtained by convolving the input map $\mathbf{x}$ with a set of $m$ pre-defined and fixed convolutional filters $\mathbf{B} \in \mathds{R}^{m \times p\times h \times w}$, followed by a non-linear activation. The corresponding output feature map value $d'$ for LBCNN is obtained by linearly combining the $m$ intermediate bit maps via convolution with $m$ convolutional filters with parameters: $v_1, v_2, \ldots, v_m$ of size $1\times1$. This entire process can be expressed as:
\begin{align}
d' = \sigma_{\mathrm{sigmoid}} ( \underbrace{\mathbf{B}\mathbf{x}}_{m\times1} )^\top \underbrace{\mathbf{v}}_{m\times1} = \mathbf{c}_{\mathrm{sigmoid}}^\top\mathbf{v}
\label{eq:d_prime}
\end{align}
where $\mathbf{B}$ is now a 2D matrix of size $m \times (p\cdot h\cdot w)$ with $m$ filters stacked as rows, with a slight abuse of notation. $\mathbf{v} = [v_1,\ldots
,v_m]^\top \in \mathds{R}^{m\times1}$.
The ReLU activation in Eq. \ref{eq:d} constraints the range of output, \ie, $d\geq0$. Eq. \ref{eq:d_prime} also places similar constraints on the output value \ie,  $\mathbf{c}_{\mathrm{sigmoid}} = \sigma_{\mathrm{sigmoid}} ( \mathbf{B}\mathbf{x} )$ $\in (0,1)$, due to the sigmoid activation. Therefore, one can always obtain a $\mathbf{v}$ such that $\mathbf{c}_{\mathrm{sigmoid}}^\top \mathbf{v} = d' = d$.

However, choosing ReLU as the LBC's activation function induces the following expression:
\begin{align}
d' = \sigma_{\mathrm{relu}} ( \mathbf{B}\mathbf{x})^\top \mathbf{v} = \mathbf{c}_{\mathrm{relu}}^\top\mathbf{v}
\label{eq:d_prime2}
\end{align}
We consider two cases (i) $d=0$: since $\mathbf{c}_{\mathrm{relu}} = \sigma_{\mathrm{relu}} ( \mathbf{B} \mathbf{x} ) \geq 0$, a vector $\mathbf{v} \in \mathds{R}^{m\times1}$ always exists such that $d'=d$. However, when (ii) $d>0$: it is obvious that the approximation does not hold when $\mathbf{c}_{\mathrm{relu}}=\mathbf{0}$. Next we will show the conditions (Theorem \ref{th:main}) under which $\mathbf{c}_{\mathrm{relu}} > 0$ to ensure that the approximation $d'\approx d$ holds. 

\begin{definition}[subgaussian random variable]
A random variable $X$ is called subgaussian if there exist constants $\beta$, $\kappa>0$, such that $\mathbb{P}(|X|\geq t) \leq \beta e^{-\kappa t^2}$ for all $t>0$.
\end{definition}

\begin{lemma}\label{lemma:expectation}
Let $X$ be a subgaussian random variable with $\mathbb{E}[X]=0$, then there exists a constant $c$ that only depends on $\beta$ and $\kappa>0$ such that $\mathbb{E}[ \exp( \theta X )  ] \leq \exp(c\theta^2)$ for all $\theta \in \mathds{R}$.
Conversely, if the above inequality holds, then $\mathbb{E}[X]=0$ and $X$ is subgaussian with parameters $\beta=2$ and $\kappa = 1/(4c)$.
\end{lemma}

\begin{definition}[isotropic random vector]
Let $\boldsymbol{\epsilon}$ be a random vector on $\mathds{R}^N$. If $\mathbb{E}[ |\langle \boldsymbol{\epsilon}, \mathbf{x} \rangle|^2 ] = \| \mathbf{x} \|_2^2$ for all $\mathbf{x}\in \mathds{R}^{N}$, then $\boldsymbol{\epsilon}$ is called an isotropic random vector.
\end{definition}

\begin{definition}[subgaussian random vector]
Let $\boldsymbol{\epsilon}$ be a random vector on $\mathds{R}^N$. If for all $\mathbf{x}\in \mathds{R}^{N}$ with $\| \mathbf{x} \|_2 = 1$, the random variable $\langle \boldsymbol{\epsilon}, \mathbf{x} \rangle$ is subgaussian with subgaussian parameter $c$ being independent of $\mathbf{x}$, that is
\begin{align}
\mathbb{E}[ \exp( \theta \langle \boldsymbol{\epsilon}, \mathbf{x} \rangle )  ] \leq \exp(c\theta^2),\mathrm{~for~all~}\theta \in \mathds{R}, ~\| \mathbf{x} \|=1
\label{eq:9.5}
\end{align}
then $\boldsymbol{\epsilon}$ is called a subgaussian random vector.
\end{definition}

\begin{lemma}\label{lemma:subgaussian}
Bernoulli random matrices are subgaussian matrices.
\end{lemma}

\begin{lemma}\label{lemma:isotropic}
Bernoulli random vectors are isotropic.
\end{lemma}

\begin{lemma}\label{lemma:concentration}
Let $\mathbf{B}$ be an $m \times N$ random matrix with independent, isotropic, and subgaussian rows with the same subgaussian parameter $c$ in (\ref{eq:9.5}). Then, for all $\mathbf{x}\in\mathds{R}^N$ and every $t\in (0,1)$,
\begin{align}
\mathbb{P} \left( \left| \frac{1}{m} \| \mathbf{Bx} \|_2^2 - \| \mathbf{x} \|_2^2 \right| \geq t \| \mathbf{x} \|_2^2 \right) \leq 2 \exp(-\tilde{c} t^2m)
\end{align}
where $\tilde{c}$ only depends on $c$.
\end{lemma}

\begin{theorem}[]\label{th:main}
Let $\mathbf{B} \in \mathds{R}^{ m \times N}$ be a Bernoulli random matrix with the same subgaussian parameter $c$ in (\ref{eq:9.5}), and $\mathbf{x}\in\mathds{R}^{N}$ be a fixed vector and $\|\mathbf{x}\|_2>0$, with $N = p\cdot h\cdot w$. Let $\boldsymbol{\xi} = \mathbf{Bx}\in\mathds{R}^{m} $. Then, for all $t\in(0,1)$, there exists a matrix $\mathbf{B}$ and an index $i\in[m]$ such that
\begin{align}
\mathbb{P} \left( \xi_i \geq \underbrace{\sqrt{(1-t)} \| \mathbf{x} \|_2}_{>0}   \right) \geq 1 - 2 \exp(-\tilde{c} t^2m)
\end{align}
\end{theorem}

\noindent Theorem~\ref{th:main} shows that with high probability, elements in the $\boldsymbol{\xi}=\mathbf{Bx}$ vector are greater than zero, which ensures that for the case when $d>0$ under ReLU activation, there is a vector $\mathbf{v}$ such that $d \approx d'$ with high probability. 

This analysis is valid for a single image patch that is convolved with CNN and LBCNN filters. We now consider a relaxed scenario with a total of $\tau$ patches per image. The output feature map for the image is a $\tau$ dimensional vector $\mathbf{d} \in \mathds{R}^\tau$ with each element $d_i, i\in[\tau]$ being the scalar output for $i$-th patch in the CNN. Similarly, for LBCNN the output feature map is a vector $\mathbf{d}' = \mathbf{C}_{\mathrm{relu}}^\top\mathbf{v}$,
where $\mathbf{C}_{\mathrm{relu}}\in\mathds{R}^{m\times \tau}$ and each column in $\mathbf{C}_{\mathrm{relu}}$ corresponds to the $m$ bit maps from each of the $\tau$ image patches. Observe that vector $\mathbf{v}$ is now shared across all the $\tau$ image patches \ie, the $\tau$ columns in $\mathbf{C}_{\mathrm{relu}}$ to approximate $\mathbf{d}$. When $\tau\leq m$, a vector $\mathbf{v}$ can be solved for such that $\mathbf{d}' = \mathbf{C}_{\mathrm{relu}}^\top\mathbf{v}$. However, when $\tau > m$, the problem reduces to an over-determined system of linear equations and a least-square error solution $\tilde{\mathbf{v}}$ is given by $\tilde{\mathbf{v}} = (\mathbf{C}\mathbf{C}^\top)^{-1}\mathbf{C}\mathbf{d}'$, such that $\mathbf{d}' \approx \mathbf{C}_{\mathrm{relu}}^\top \tilde{\mathbf{v}}$. This analysis suggests that using a larger number of intermediate filters $m$ can result in a better approximation of the standard convolutional layer.

\begin{figure}
  \centering
  \includegraphics[width=0.225\textwidth]{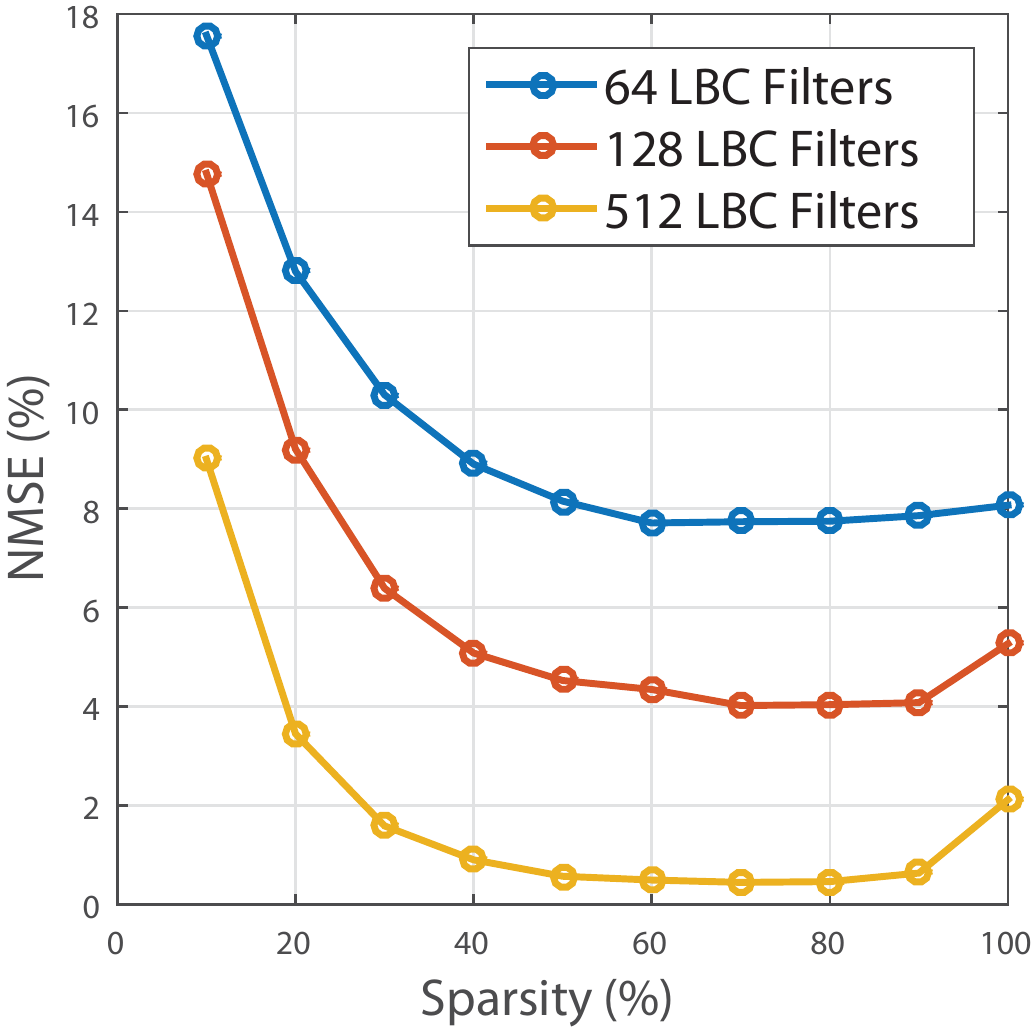}
  \includegraphics[width=0.23\textwidth]{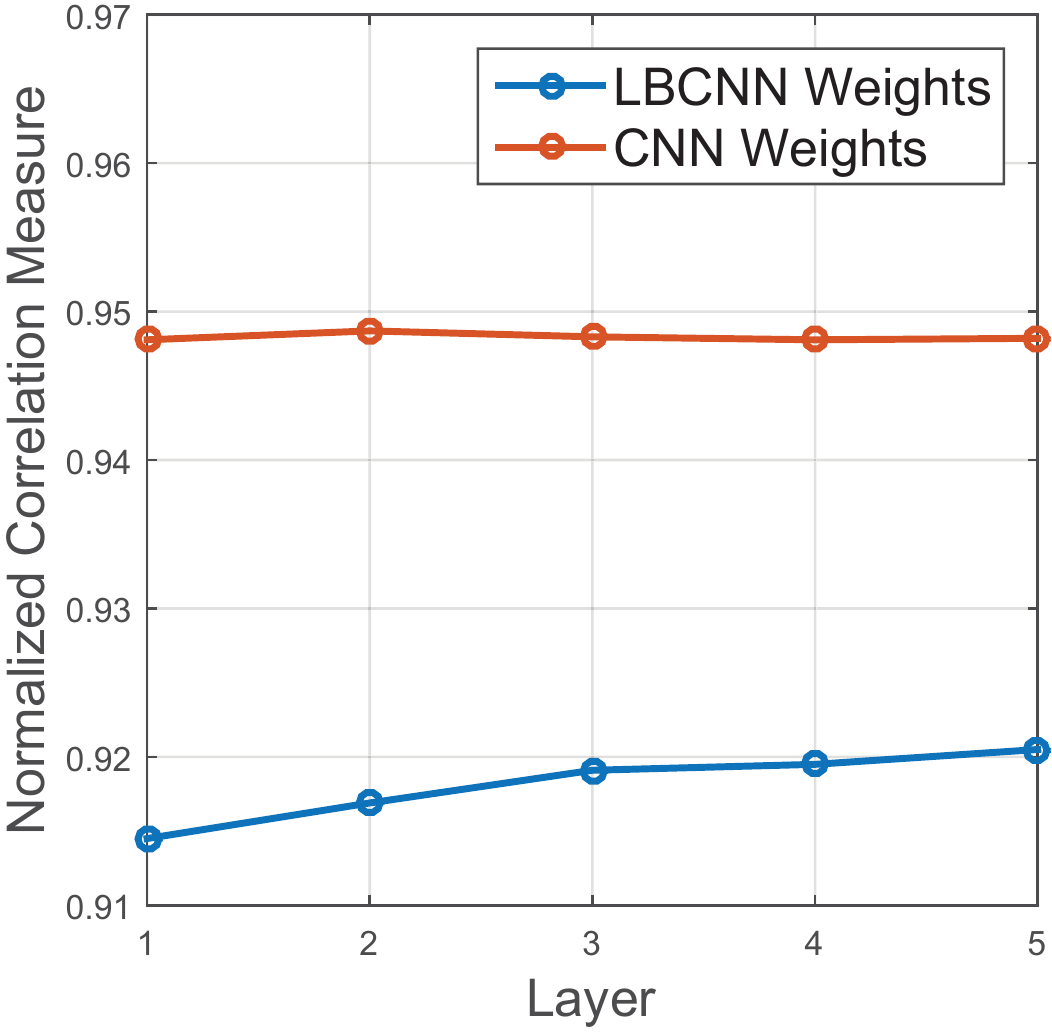}
  \caption{(L) NMSE between $\mathbf{d}'$ and $\mathbf{d}$ with increasing levels of sparsity within the LBC filters. (R) Normalized correlation measure for LBCNN and CNN filters. The smaller the value, the more de-correlated they are.}
  \label{fig:decorr_mse}
\end{figure}

Empirically we can measure how far $\mathbf{d}'$ is from $\mathbf{d}$ by measuring the normalized mean square error (NMSE): $\| \mathbf{d}'-\mathbf{d} \|_2^2 / \| \mathbf{d} \|_2^2$. We take the entire 50,000 $32\times 32$ images from CIFAR-10 training set and measure the NMSE, as shown in Figure~\ref{fig:decorr_mse} (L). For the CNN, dense real-valued filters are independently generated as Gaussian random filters, for each individual image. For the LBCNN, the sparse LBC filters are also independently generated for each individual image. Experiments are repeated for 10 levels of sparsity $(10\%, 20\%, \ldots, 100\%)$ and 3 choices of number of intermediate channels, 64, 128 and 512. We can see that the approximation is better using more filters, and with higher sparsity, with the exception of sparsity being 100\%. We conjecture that this may be due to that fact that $\mathbf{d}$ is actually sparse, due to ReLU activation, and therefore enforcing no sparsity constraints on the LBC filters $\mathbf{B}$ actually makes the approximation harder.

\section{Experimental Results}\label{sec:exp}

We will evaluate the efficacy of the proposed LBC layer and compare its performance to a standard convolutional layer on several datasets, both small scale and large scale. 

\vspace{1mm}
\noindent\textbf{Datasets:} We consider classification tasks on four different visual datasets, MNIST, SVHN, CIFAR-10, and ILSVRC-2012 ImageNet classification challenge. The MNIST \cite{mnist} dataset is composed of a training set of 60K and a testing set of 10K $32\times32$ gray-scale images of hand-written digits from 0 to 9. SVHN \cite{svhn} is also a widely used dataset for classifying digits, house number digits from street view images in this case. It consists of a training set of 604K and a testing set of 26K $32\times32$ color images showing house number digits. CIFAR-10 \cite{cifar} is an image classification dataset containing a training set of 50K and a testing set of 10K $32\times32$ color images across the following 10 classes: airplanes, automobiles, birds, cats, deers, dogs, frogs, horses, ships, and trucks. \textcolor{black}{The ImageNet ILSVRC-2012 classification dataset \cite{imagenet2012} consists of 1000 classes, with 1.28 million images for training and 50K images for validation. We first consider a subset of this dataset. We randomly selected 100 classes with the largest number of images (1300 training images in each class, for a total of 130K training images and 5K testing images.), and report top-1 accuracy on this subset. Full ImageNet experimental results are also reported in the subsequent section.}

\vspace{1mm}
\noindent\textbf{Implementation Details:} Conceptually LBCNN can be easily implemented in any existing deep learning framework. Since the convolutional weights are fixed, we do not have to compute the gradients nor update the weights. This leads to savings both from a computational point of view and memory as well. Furthermore, since the weights are binary the convolution operation can be performed purely through additions and subtractions. We base the model architectures we evaluate in this paper on ResNet \cite{identityMapping}, with default $3\times3$ filter size. Our basic module is the LBC module shown in Figure~\ref{fig:pipeline} along with an identity connection as in ResNet. We experiment with different numbers of LBC units, 10, 20 and 75, which is equivalent to 20, 40, and 150 convolutional layers. For LBCNN the convolutional weights are generated following the procedure described in Section~\ref{sec:train}. We use 512 randomly generated anchor weights, with a sparsity of 0.1, 0.5 or 0.9, for all of our experiments. Spatial average pooling is adopted after the convolution layers to reduce the spatial dimensions of the image to $6\times6$. We use a learning rate of 1e-3 and adopt the learning rate decay schedule from \cite{identityMapping}. We use ReLU instead of sigmoid as our non-linear function for computational efficiency and faster convergence. An important and practical consideration is to avoid using a ReLU activation just prior to the LBC layer. This is to ensure that there is no irrecoverable loss of information due to the sparsity in both the input (due to ReLU activation) and the convolutional weights.


\vspace{1mm}
\noindent\textbf{Baselines:} To ensure a fair comparison and to quantify the exact empirical difference between our LBCNN approach and a traditional CNN, we use the exact same architecture for both the networks, albeit with sparse, binary and fixed weights in LBCNN and dense learnable weights for CNN. We also use the exact same data and hyper-parameters in terms of the number of convolutional filters, initial learning rate and the learning rate schedule. Consequently in these experiments with $3\times 3$ convolutional kernels, LBCNN has $10\times$ fewer learnable parameters (the baseline CNN also includes a $1\times1$ convolutional layer).
\begin{table}
\centering
\tiny
\begin{tabular}{cccccccccc}
\toprule
  $q$ & 16 & 32 & 64 & 128 & 192 & 256 & 384 & 512 \\ 
  \midrule
  \textbf{LBCNN}      & 82.74 & 85.57 & 88.18 & 90.70 & 91.58 & 92.13 & \textbf{92.96} & 92.09 \\
  \textbf{LBCNN-share}& 82.70 & 85.26 & 87.85 & 90.26 & 91.37 & 91.72 & \textbf{92.91} & 91.83 \\
  Baseline            & 84.13 & 86.30 & 88.77 & 90.86 & 91.69 & 92.15 & \textbf{92.93} & 91.87 \\
\bottomrule
\end{tabular}
\caption{Classification accuracy $(\%)$ on CIFAR-10 with 20 convolution layers and 512 LBC filters on LBCNN, LBCNN-share, and CNN baseline.}
\label{tab:cifar}
\end{table}

\vspace{1mm}
\noindent\textbf{Results on MNIST, SVHN, and CIFAR-10:} 
Table~\ref{tab:cifar} compares the accuracy achieved by LBCNN, LBCNN with shared convolutional weights and the corresponding network with a regular convolutional layer on the CIFAR-10 dataset. Note that with a fixed number of convolutional layers, number of input and output channels, performance of the networks increases with an increase in the number of output channels $q$. Significantly, LBCNN with $10\times$ fewer parameters performs as well as the corresponding CNN.

%

Table~\ref{tab:result} consolidates the images classification results from our experiments on various datasets. The best performing LBCNNs are compared to their corresponding CNN baseline, as well as to state-of-the-art methods such as BinaryConnect \cite{binaryConnect}, Binarized Neural Networks (BNN) \cite{binaryNet}, ResNet \cite{resNet}, Maxout Network \cite{maxout}, Network in Network (NIN) \cite{nin}. For each dataset under consideration the best performing LBCNN models are:
\begin{itemize}
    \item MNIST: 150 convolutional layers (75 LBCNN modules), 512 LBC filters, 16 output channels, 0.5 sparsity, 128 hidden units in the fully connected layer.
    \item SVHN: 80 convolutional layers (40 LBCNN modules), 512 LBC filters, 16 output channels, 0.9 sparsity, 512 hidden units in the fully connected layer.
    \item CIFAR-10: 100 convolutional layers (50 LBCNN modules), 512 LBC filters, 384 output channels, 0.1 sparsity, 512 hidden units in the fully connected layer.
\end{itemize}




\begin{figure}
  \centering
  \includegraphics[width=0.22\textwidth]{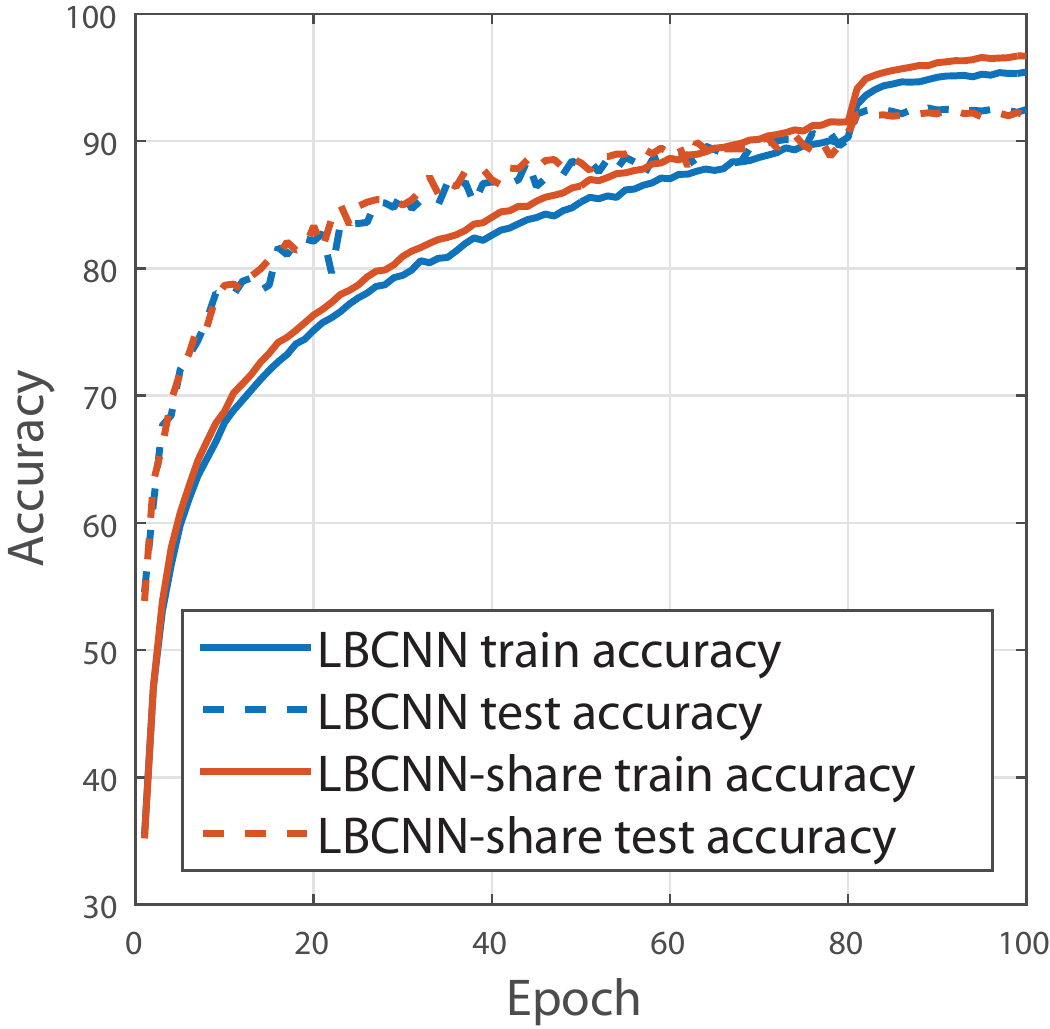}
  \includegraphics[width=0.24\textwidth]{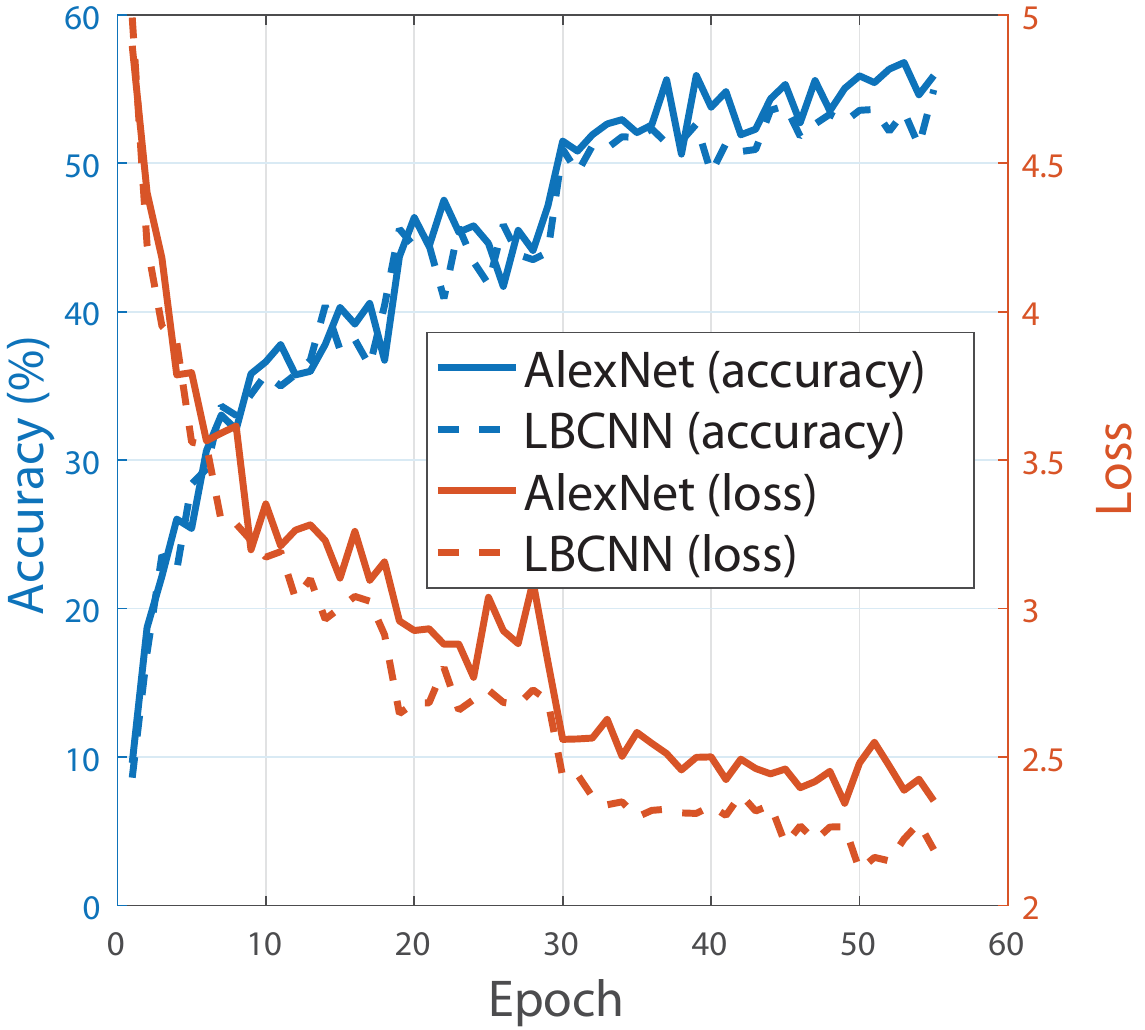}
  \caption{(L) Accuracy of the best performing LBCNN and LBCNN-share on CIFAR-10. (R) Accuracy and loss on full ImageNet classification.}
  \label{fig:dup_alexnet}
\end{figure}

\begin{table*}
\centering
\scriptsize
\begin{tabular}{cccccccc}
\toprule
           & \textbf{LBCNN} & Baseline & BinaryConnect \cite{binaryConnect} & BNN \cite{binaryNet,qnn} & ResNet \cite{resNet} & Maxout \cite{maxout} & NIN \cite{nin}\\ 
  \midrule
  MNIST    & 99.51 & 99.48 & 98.99 & 98.60  & /     & 99.55 & 99.53 \\
  SVHN     & 94.50 & 95.21 & 97.85 & 97.49  & /     & 97.53 & 97.65 \\
  CIFAR-10 & 92.99 \emph{(93.66 NetEverest}) & 92.95 & 91.73 & 89.85  & 93.57 & 90.65 & 91.19 \\
\bottomrule
\end{tabular}
\caption{Classification accuracy $(\%)$. LBCNN column only shows the best performing model and the Baseline column shows the particular CNN counterpart.}
\label{tab:result}
\end{table*}

\vspace{1mm}
\noindent\textbf{LBCNN with Shared Weights:} 
We consider a scenario where all the LBC layers in the network share the same set of convolutional weights, as opposed to randomly generating new convolutional weights at each layer. For a network with $D$ LBC layers sharing the convolutional weights across the layers results in a model size that is roughly smaller by a factor of $D$. As can be seen from the second row in Table~\ref{tab:cifar} and in Figure~\ref{fig:dup_alexnet} (L), the performance of the network with weight sharing is comparable to a network without weight sharing. This experiment demonstrates the practicality of using a LBCNN on memory constrained embedded systems.




\vspace{1mm}
\noindent\textbf{NetEverest:}
With at least 9$\times$ parameter reduction, one can now train much deeper networks, going roughly from 100 to 1000 layers, or from 1000 to 10000 layers. The LBC module allows us to train extremely deep CNN efficiently with 8848 convolutional layers (4424 LBC modules), dubbed \textit{NetEverest}, using a single nVidia Titan X GPU. The architecture of NetEverest: 8848 convolutional layers (4424 LBC modules), 32 LBC filters, 32 output channels, 0.1 sparsity, 512 hidden units in the fully connected layer. This network achieves the highest accuracy on CIFAR-10 among our experiments as shown in Table~\ref{tab:result}.

\vspace{1mm}
\noindent\textbf{Results on 100-Class ImageNet Subset:} 
We report the top-1 accuracy on a 100-Class subset of ImageNet 2012 classification challenge dataset in Table~\ref{tab:imagenet}. The input images of ImageNet are of a much higher resolution than those in MNIST, SVHN, and CIFAR-10, allowing us to experiments with the different LBC filter sizes. Both LBCNN and our baseline CNN share the same architecture: 48 convolutional layers (24 LBC modules), 512 LBC filters, 512 output channels, 0.9 sparsity, 4096 hidden units in the fully connected layer.

\begin{table}
\centering
\scriptsize
\begin{tabular}{cccccccc}
\toprule
  LBC Filter Size & 3$\times$3 & 5$\times$5 & 7$\times$7 & 9$\times$9 & 11$\times$11 & 13$\times$13  \\ 
  \midrule
  \textbf{LBCNN}& 62.56 & 62.29 & 62.80 & 63.24 & 63.08 & 62.43  \\
  Baseline      & 65.74 & 64.90 & 66.53 & 65.91 & 65.22 & 64.94  \\
  \bottomrule
\end{tabular}
\caption{Classification accuracy $(\%)$ on 100-class ImageNet with varying LBC filter sizes.}
\label{tab:imagenet}
\end{table}

\begin{table}
\centering
\scriptsize
\begin{tabular}{ccc}
\toprule
  Layers & AlexNet \cite{alexNet} & \textbf{LBCNN (AlexNet)}  \\ 
  \midrule
  Layer 1 & $96\times(11\times11\times3)=34,848$   & $96\times256=24,576$  \\
  Layer 2 & $256\times(5\times5\times48)=307,200$  & $256\times256=65,536$ \\
  Layer 3 & $384\times(3\times3\times256)=884,736$ & $384\times256=98,304$ \\
  Layer 4 & $384\times(3\times3\times192)=663,552$ & $384\times256=98,304$ \\
  Layer 5 & $256\times(3\times3\times192)=442,368$ & $256\times256=65,536$ \\
  \midrule
  Total   & $2,332,704~(\sim2.33M)$   &  $352,256~(\sim0.352M)$\\
  \bottomrule
\end{tabular}
\caption{Comparison of the number of learnable parameters in convolutional layers in AlexNet and AlexNet with LBCNN modules. The proposed method saves $6.622\times$ learnable parameters in the convolutional layers.}
\label{tab:alexnetweight}
\end{table}

\vspace{1mm}
\noindent\textbf{Results on Full ImageNet:} 
We train a LBCNN version of the AlexNet \cite{alexNet} architecture on the full ImageNet classification dataset. The AlexNet architecture is comprised of five consecutive convolutional layers, and two fully connected layers, mapping the image $(224\times224\times3)$ to a 1000-dimension feature representation for classification. The number of convolutional filters used and their spatial sizes are tabulated in Table~\ref{tab:alexnetweight}. For this experiment, we create a LBCNN version of the AlexNet architecture by replacing each convolutional layer in AlexNet with a LBC layer with the same number input and output channels and size of filter. Table~\ref{tab:alexnetweight} compares the number of learnable parameters in convolutional layers in both AlexNet and its LBCNN version by setting the number of output channels to $q=256$. As can be seen, LBCNN acheives a $6.622\times$ reduction in the number of learnable parameters in the convolutional layers while performing comparably to AlexNet (see Table~\ref{tab:imagenet_full}). The progression in the validation accuracy and training loss of AlexNet and its corresponding LBCNN version set for 55 epochs is shown in Figure~\ref{fig:dup_alexnet}.

\begin{table}
\centering
\scriptsize
\begin{tabular}{cccc}
\toprule
  ~ & \textbf{LBCNN} & AlexNet (ours) & AlexNet (BLVC) \cite{blvcAlexnet} \\ 
  \midrule
  ImageNet  & 54.9454 & 56.7821 & 56.9 \\
  \bottomrule
\end{tabular}
\caption{Classification accuracy $(\%)$ on full ImageNet.}
\label{tab:imagenet_full}
\end{table}

\begin{figure*}
\centering
\includegraphics[width=0.245\textwidth]{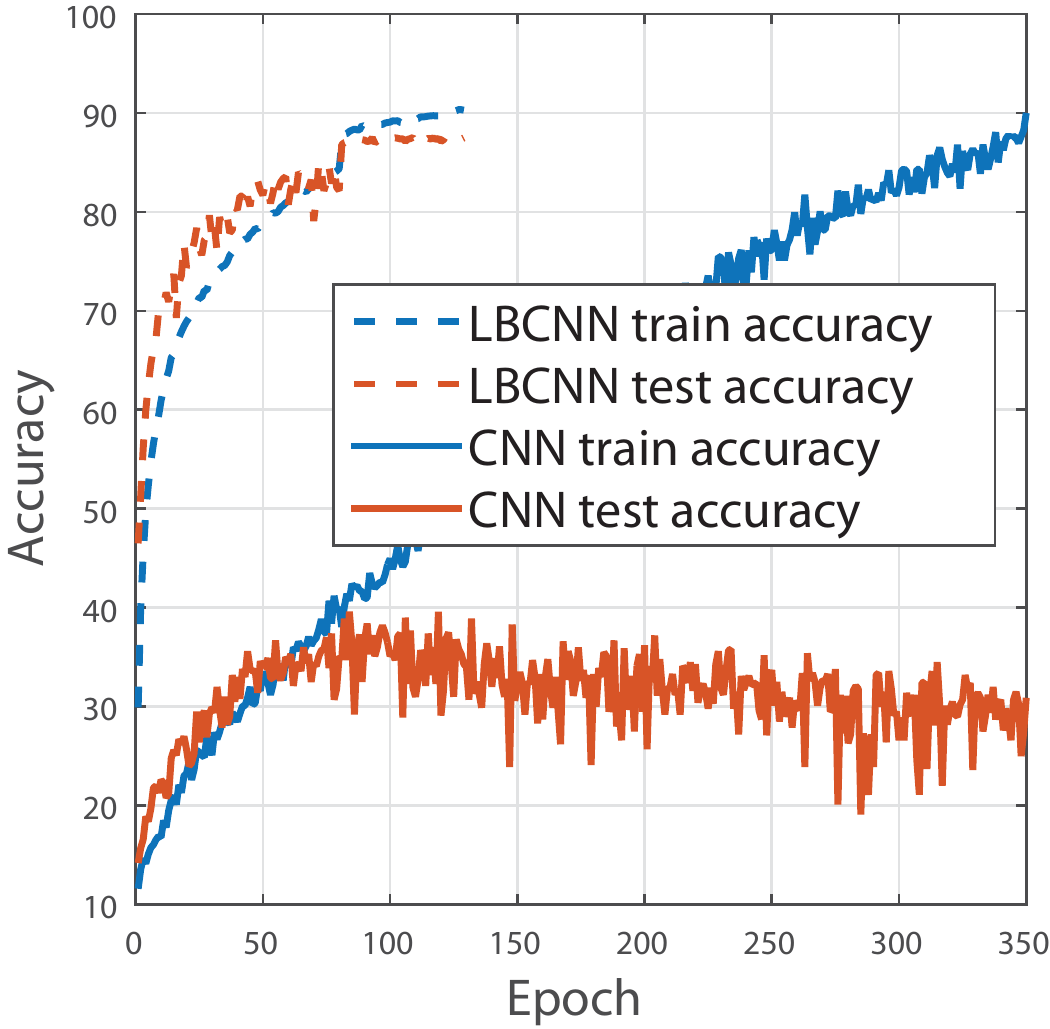}
\includegraphics[width=0.245\linewidth]{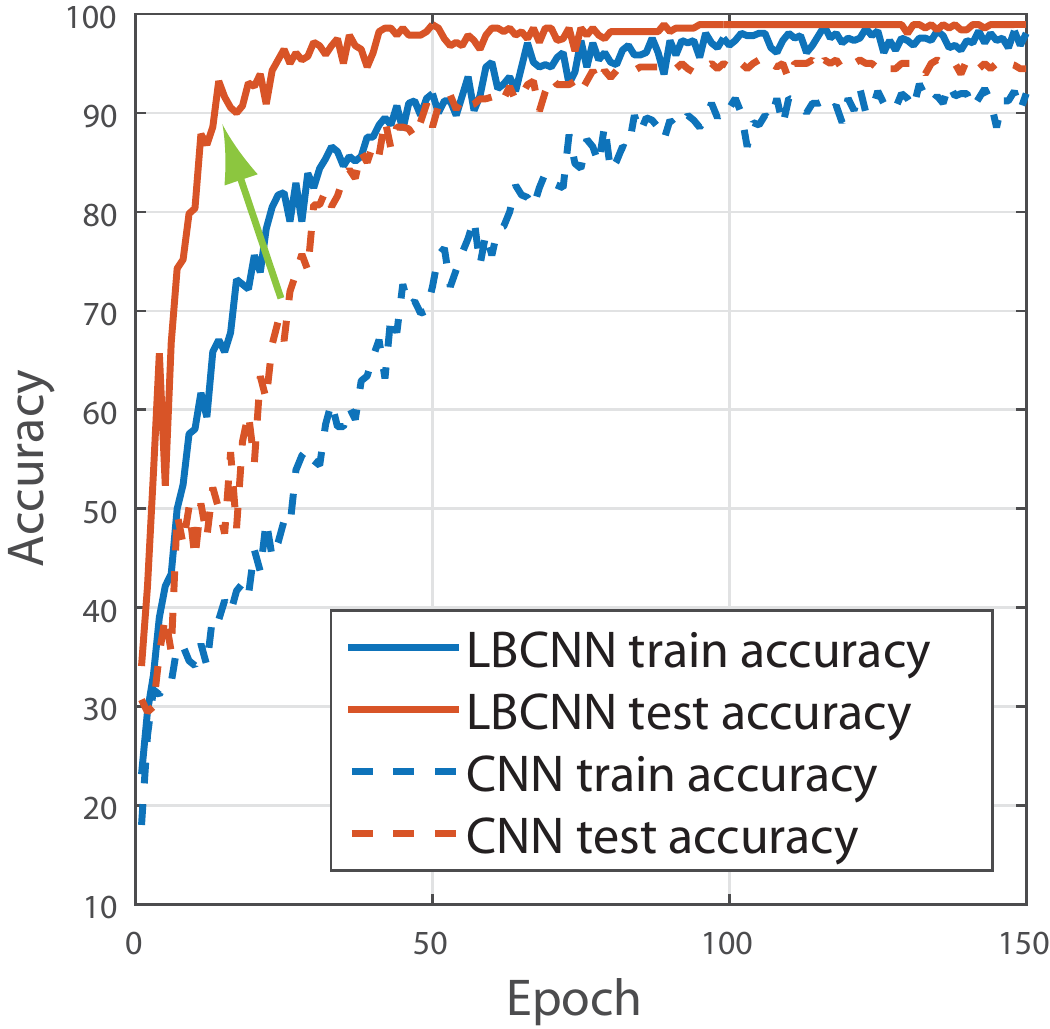}
\includegraphics[width=0.245\linewidth]{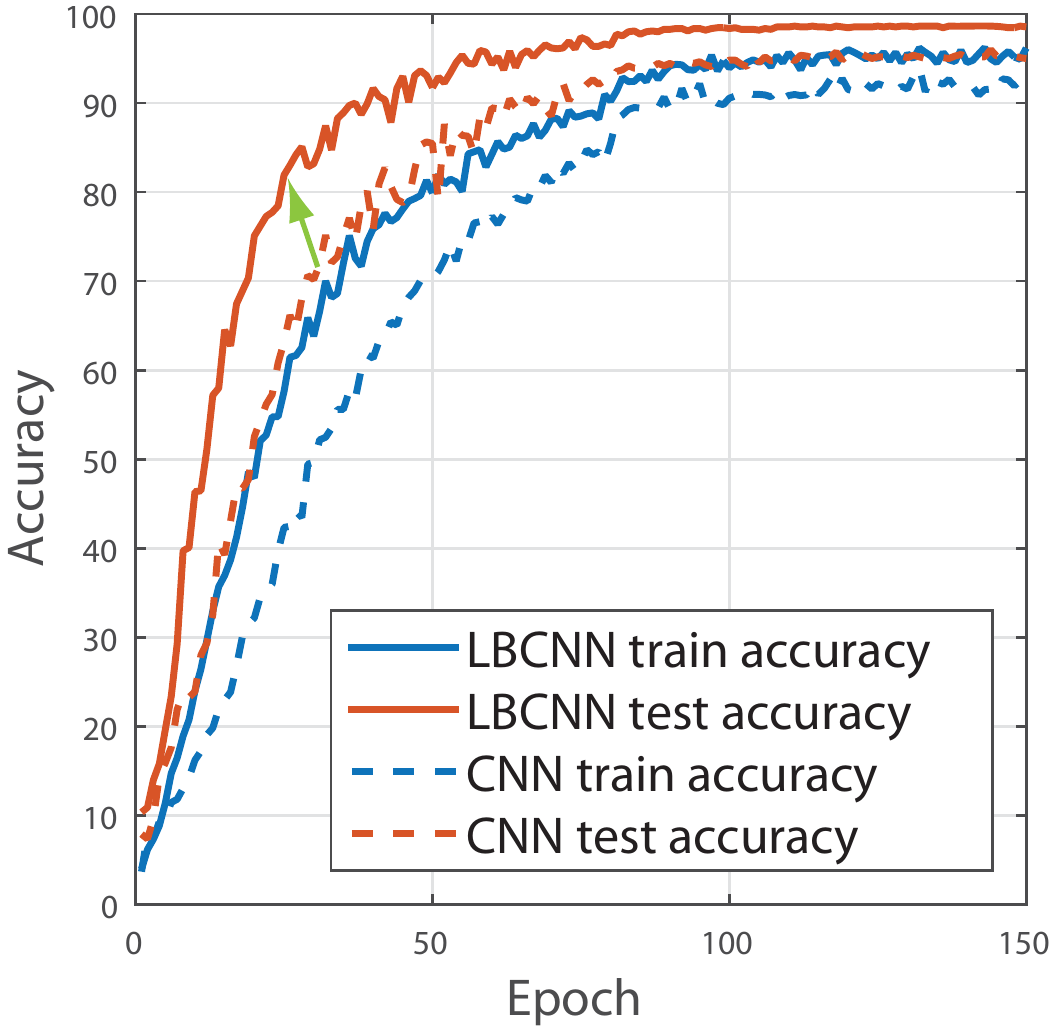}
\includegraphics[width=0.245\linewidth]{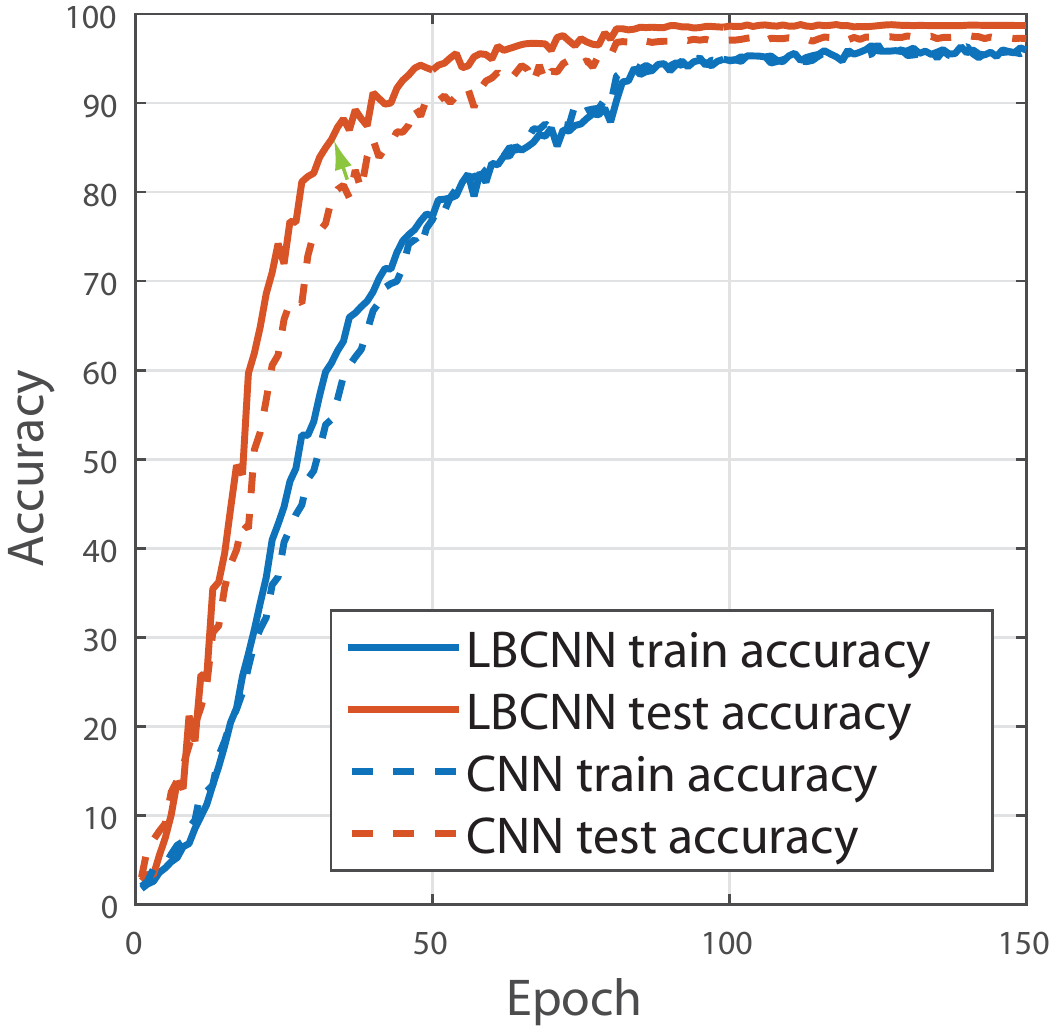}
\caption{(L1) Results on overfitting experiments. (R3) Results on the FRGC 10-class, 50-class, and 100-class experiments respectively.}
\label{fig:over-fitting-three-plots}
\end{figure*}

\section{Discussion}

\noindent We now discuss some computational and statistical advantages afforded by the proposed local binary convolution layer over a regular convolutional layer.




\vspace{1mm}
\noindent\textbf{Computational:} The parametrization of the LBC layer reduces the number of learnable parameters by a factor of $9\times$ to $169\times$ during training and inference. Furthermore, the sparse and binary nature of the convolutional weights further reduces the computational complexity and memory and space requirements both during training and inference. The lower memory requirements enables learning of much deep neural networks thereby allowing better representations to be learned through deeper architectures \cite{vgg,resNet,identityMapping}. Furthermore, sharing the convolutional weights across all the LBC layers, leads to further reduction in memory requirements thereby enabling learning of deep CNNs on resource constrained embedded systems. 



\vspace{1mm}
\noindent\textbf{Statistical:} LBCNN, being a simpler model with fewer learnable parameters compared to a CNN, can effectively regularize the learning process and prevent over-fitting. High capacity models such as deep CNNs with a regular convolutional layer typically consists of a very large number of parameters. Methods such as Dropout \cite{dropout}, DropConnect \cite{dropConnect}, and Maxout \cite{maxout} have been introduced to regularize the fully connected layers of a network during training to avoid over-fitting. As opposed to regularizing the fully connected layers \cite{dropout,dropConnect,decorrelating} of a network, LBCNN directly regularizes the convolutional layers, which is also important as discussed in \cite{dropout,elu}. 

Network regularization techniques such as Dropout \cite{dropout} and Batch Normalization \cite{batchNorm} prevent co-adaptation of neuron activations and reduce internal co-variate shift. Recently Cogswell \etal \cite{decorrelating} propose a method to explicitly de-correlate and minimize the cross-covariance of hidden activations, to improve performance and prevent over-fitting. It encourages diverse or non-redundant representations. LBCNN naturally provides de-correlation for the activations since the convolutional filters are randomly generated sparse Bernoulli filters. Figure~\ref{fig:decorr_mse} (R) shows the amount of normalized correlation $(\| \boldsymbol{\Sigma} \|_F^2 - \| \mathrm{diag} (\boldsymbol{\Sigma}) \|_2^2)/\| \boldsymbol{\Sigma} \|_F^2$ in both LBCNN and CNN filters for the first 5 layers of the best-performing architecture on CIFAR-10 described in Section~\ref{sec:exp}. Smaller values of the normalized correlation correspond to greater de-correlation between the activations. 


\vspace{1mm}
\noindent\textbf{Sample Complexity:} The lower model complexity of LBCNN makes them an attractive option for learning with low sample complexity. To demonstrate the statistical efficiency of LBCNN we perform an experiment on a subset of the CIFAR-10 dataset. The training subset randomly picks $25\%$ images $(5000\times0.25=1250)$ per class while keeping the testing set intact. We choose the best-performing architecture on CIFAR-10 described in Section~\ref{sec:exp} for both the CNN and LBCNN. The results shown in Figure~\ref{fig:over-fitting-three-plots} (L1) demonstrates that LBCNN trains faster and is less prone to over-fitting on the training data. To provide an extended evaluation, we perform additional face recognition on the FRGC v2.0 dataset \cite{frgc} experiments under a limited sample complexity setting. The number of images in each class ranges from 6 to 132 (51.6 on average). While there are 466 classes in total, we experiment with increasing number of randomly selected classes (10, 50 and 100) with a 60-40 train/test split. Across the number of classes, our network parameters remain the same except for the classification fully connected layer at the end. We make a few observations from our findings (see Figure~\ref{fig:over-fitting-three-plots} (R3)): (1) LBCNN converges faster than CNN, especially on small datasets and (2) LBCNN outperforms CNN on this task. Lower model complexity helps LBCNN prevent over-fitting especially on small to medium-sized datasets.

\section{Conclusions}\label{sec:conc}

Motivated by traditional local binary patterns, in this paper, we proposed local binary convolution (LBC) layer as an alternative to the convolutional layers in standard CNN. The LBC layer comprises of a set of sparse, binary and randomly generated set of convolutional weights that are fixed and a set of learnable linear weights. We demonstrate, both theoretically and empirically, that the LBC module is a good approximation of a standard convolutional layer while also resulting in a significant reduction in the number of parameters to be learned at training, $9\times$ to $169\times$ for $3\times 3$ and $13\times 13$ sized filters respectively. CNNs with LBC layers are well suited for low sample complexity learning of deep CNNs in resource constrained environments due their low model and computational complexity. The proposed LBCNN demonstrates excellent performance and performs as well as standard CNNs on multiple small and large scale datasets across different network architectures.




\newpage
\clearpage

{\small
\bibliographystyle{ieee}
\bibliography{refs_lbcnn}

\begin{thebibliography}{10}\itemsep=-1pt

\bibitem{blvcAlexnet}
{Berkeley Vision and Learning Center (BLVC)}.
\newblock {BVLC AlexNet Accuracy on ImageNet 2012 Validation Set}.
\newblock {\em
  https://github.com/BVLC/caffe/wiki/Models-accuracy-on-ImageNet-2012-val},
  2015.

\bibitem{hashing}
W.~Chen, J.~T. Wilson, S.~Tyree, K.~Q. Weinberger, and Y.~Chen.
\newblock {Compressing Neural Networks with the Hashing Trick}.
\newblock In {\em 32nd International Conference on Machine Learning (ICML)},
  2015.

\bibitem{elu}
D.-A. Clevert, T.~Unterthiner, and S.~Hochreiter.
\newblock {Fast and Accurate Deep Network Learning by Exponential Linear Units
  (ELUs)}.
\newblock In {\em International Conference on Learning Representations (ICLR)},
  2016.

\bibitem{decorrelating}
M.~Cogswell, F.~Ahmed, R.~Girshick, L.~Zitnick, and D.~Batra.
\newblock {Reducing Overfitting in Deep Networks by Decorrelating
  Representations}.
\newblock In {\em International Conference on Learning Representations (ICLR)},
  2016.

\bibitem{binaryNet}
M.~Courbariaux and Y.~Bengio.
\newblock {BinaryNet: Training Deep Neural Networks with Weights and
  Activations Constrained to +1 or -1}.
\newblock {\em arXiv preprint arXiv:1602.02830}, 2016.

\bibitem{binaryConnect}
M.~Courbariaux, Y.~Bengio, and J.-P. David.
\newblock {BinaryConnect: Training Deep Neural Networks with binary weights
  during propagations}.
\newblock In {\em Advances in Neural Information Processing Systems (NIPS)},
  pages 3105--3113, 2015.

\bibitem{predicting}
M.~Denil, B.~Shakibi, L.~Dinh, M.~Ranzato, and N.~{de Freitas}.
\newblock {Predicting Parameters in Deep Learning}.
\newblock In {\em Advances in Neural Information Processing Systems (NIPS)},
  pages 2148--2156, 2013.

\bibitem{neuromorphic}
S.~K. Esser, R.~Appuswamy, P.~Merolla, J.~V. Arthur, and D.~S. Modha.
\newblock {Backpropagation for Energy-efficient Neuromorphic Computing}.
\newblock In {\em Advances in Neural Information Processing Systems (NIPS)},
  pages 1117--1125, 2015.

\bibitem{maxout}
I.~J. Goodfellow, D.~Warde-Farley, M.~Mirza, A.~Courville, and Y.~Bengio.
\newblock {Maxout Networks}.
\newblock In {\em 30th International Conference on Machine Learning (ICML)},
  2013.

\bibitem{deepCompression}
S.~Han, H.~Mao, and W.~J. Dally.
\newblock {Deep Compression: Compressing Deep Neural Networks with Pruning,
  Trained Quantization and Huffman Coding}.
\newblock In {\em International Conference on Learning Representations (ICLR)},
  2016.

\bibitem{efficient}
S.~Han, J.~Pool, J.~Tran, and W.~Dally.
\newblock {Learning both Weights and Connections for Efficient Neural Network}.
\newblock In {\em Advances in Neural Information Processing Systems (NIPS)},
  pages 1135--1143, 2015.

\bibitem{resNet}
K.~He, X.~Zhang, S.~Ren, and J.~Sun.
\newblock {Deep Residual Learning for Image Recognition}.
\newblock In {\em IEEE International Conference on Computer Vision and Pattern
  Recognition (CVPR)}, pages 770--778, 2016.

\bibitem{identityMapping}
K.~He, X.~Zhang, S.~Ren, and J.~Sun.
\newblock {Identity Mappings in Deep Residual Networks}.
\newblock In {\em European Conference on Computer Vision (ECCV)}, pages
  630--645, 2016.

\bibitem{qnn}
I.~Hubara, M.~Courbariaux, D.~Soudry, R.~El-Yaniv, and Y.~Bengio.
\newblock Quantized neural networks: Training neural networks with low
  precision weights and activations.
\newblock {\em arXiv preprint arXiv:1609.07061}, 2016.

\bibitem{squeezeNet}
F.~N. Iandola, M.~W. Moskewicz, K.~Ashraf, S.~Han, W.~J. Dally, and K.~Keutzer.
\newblock {SqueezeNet: AlexNet-level Accuracy with 50x Fewer Parameters and
  $<$1MB Model Size}.
\newblock {\em arXiv preprint arXiv:1602.07360}, 2016.

\bibitem{batchNorm}
S.~Ioffe and C.~Szegedy.
\newblock {Batch Normalization: Accelerating Deep Network Training by Reducing
  Internal Covariate Shift}.
\newblock In {\em 32nd International Conference on Machine Learning (ICML)},
  2015.

\bibitem{Felix_tip15_spartans}
F.~Juefei-Xu, K.~Luu, and M.~Savvides.
\newblock {Spartans: Single-sample Periocular-based Alignment-robust
  Recognition Technique Applied to Non-frontal Scenarios}.
\newblock {\em IEEE Trans. on Image Processing}, 24(12):4780--4795, Dec 2015.

\bibitem{Felix_tip14_lbp}
F.~Juefei-Xu and M.~Savvides.
\newblock {Subspace-Based Discrete Transform Encoded Local Binary Patterns
  Representations for Robust Periocular Matching on NIST's Face Recognition
  Grand Challenge}.
\newblock {\em IEEE Trans. on Image Processing}, 23(8):3490--3505, Aug 2014.

\bibitem{Felix_bmvc16_invert}
F.~Juefei-Xu and M.~Savvides.
\newblock {Learning to Invert Local Binary Patterns}.
\newblock In {\em 27th British Machine Vision Conference (BMVC)}, Sept 2016.

\bibitem{cifar}
A.~Krizhevsky and G.~Hinton.
\newblock {Learning Multiple Layers of Features from Tiny Images}.
\newblock 2009.

\bibitem{alexNet}
A.~Krizhevsky, I.~Sutskever, and G.~E. Hinton.
\newblock {ImageNet Classification with Deep Convolutional Neural Networks}.
\newblock In {\em Advances in Neural Information Processing Systems (NIPS)},
  pages 1097--1105, 2012.

\bibitem{mnist}
Y.~LeCun, L.~Bottou, Y.~Bengio, and P.~Haffner.
\newblock {Gradient-based Learning Applied to Document Recognition}.
\newblock {\em Proceedings of the IEEE}, 86(11):2278--2324, 1998.

\bibitem{nin}
M.~Lin, Q.~Chen, and S.~Yan.
\newblock {Network in Network}.
\newblock In {\em International Conference on Learning Representations (ICLR)},
  2014.

\bibitem{svhn}
Y.~Netzer, T.~Wang, A.~Coates, A.~Bissacco, B.~Wu, and A.~Y. Ng.
\newblock {Reading Digits in Natural Images with Unsupervised Feature
  Learning}.
\newblock In {\em NIPS Workshop on Deep Learning and Unsupervised Feature
  Learning}, 2011.

\bibitem{c13}
T.~Ojala, M.~Pietik\"{a}inen, and D.~Harwood.
\newblock {A Comparative Study of Texture Measures with Classification Based on
  Featured Distributions}.
\newblock {\em Pattern Recognition}, 29(1):51--59, 1996.

\bibitem{frgc}
P.~J. Phillips, P.~J. Flynn, T.~Scruggs, K.~W. Bowyer, J.~Chang, K.~Hoffman,
  J.~Marques, J.~Min, and W.~Worek.
\newblock {Overview of the Face Recognition Grand Challenge}.
\newblock In {\em IEEE Conference on Computer Vision and Pattern Recognition
  (CVPR)}, volume~1, pages 947--954, 2005.

\bibitem{LBPbook}
M.~Pietik{\"a}inen, A.~Hadid, G.~Zhao, and T.~Ahonen.
\newblock {\em {Computer Vision Using Local Binary Patterns}}.
\newblock Springer, 2011.

\bibitem{xnorNet}
M.~Rastegari, V.~Ordonez, J.~Redmon, and A.~Farhadi.
\newblock {XNOR-Net: ImageNet Classification Using Binary Convolutional Neural
  Networks}.
\newblock In {\em European Conference on Computer Vision (ECCV)}, pages
  525--542, 2016.

\bibitem{imagenet2012}
O.~Russakovsky, J.~Deng, H.~Su, J.~Krause, S.~Satheesh, S.~Ma, Z.~Huang,
  A.~Karpathy, A.~Khosla, M.~Bernstein, et~al.
\newblock {ImageNet Large Scale Visual Recognition Challenge}.
\newblock {\em International Journal of Computer Vision (IJCV)},
  115(3):211--252, 2015.

\bibitem{vgg}
K.~Simonyan and A.~Zisserman.
\newblock {Very Deep Convolutional Networks for Large-scale Image Recognition}.
\newblock In {\em International Conference on Learning Representations (ICLR)},
  2015.

\bibitem{expectationBackprop}
D.~Soudry, I.~Hubara, and R.~Meir.
\newblock {Expectation Backpropagation: Parameter-free Training of Multilayer
  Neural Networks with Continuous or Discrete Weights}.
\newblock In {\em Advances in Neural Information Processing Systems (NIPS)},
  pages 963--971, 2014.

\bibitem{dropout}
N.~Srivastava, G.~Hinton, A.~Krizhevsky, I.~Sutskever, and R.~Salakhutdinov.
\newblock {Dropout: A Simple Way to Prevent Neural Networks from Overfitting}.
\newblock {\em The Journal of Machine Learning Research (JMLR)},
  15(1):1929--1958, 2014.

\bibitem{szegedy2015going}
C.~Szegedy, W.~Liu, Y.~Jia, P.~Sermanet, S.~Reed, D.~Anguelov, D.~Erhan,
  V.~Vanhoucke, and A.~Rabinovich.
\newblock {Going Deeper with Convolutions}.
\newblock In {\em IEEE Conference on Computer Vision and Pattern Recognition
  (CVPR)}, pages 1--9, 2015.

\bibitem{dropConnect}
L.~Wan, M.~Zeiler, S.~Zhang, Y.~L. Cun, and R.~Fergus.
\newblock {Regularization of Neural Networks Using Dropconnect}.
\newblock In {\em 30th International Conference on Machine Learning (ICML)},
  pages 1058--1066, 2013.

\bibitem{quantizedMobile}
J.~Wu, C.~Leng, Y.~Wang, Q.~Hu, and J.~Cheng.
\newblock {Quantized Convolutional Neural Networks for Mobile Devices}.
\newblock In {\em IEEE Conference on Computer Vision and Pattern Recognition
  (CVPR)}, June 2016.

\end{thebibliography}
}

\end{document}